\documentclass[10pt,twocolumn,letterpaper]{article}

\usepackage[pagenumbers]{cvpr} 

\usepackage[pagebackref,breaklinks,colorlinks,citecolor=cvprblue]{hyperref}
\usepackage{amsmath}
\usepackage{amssymb}
\usepackage{booktabs}
\usepackage{float, amsfonts, amsmath, amssymb, multirow, verbatim,array, bm}
\usepackage{threeparttable}
\usepackage{algorithm, algpseudocode} 
\usepackage{xcolor}
\usepackage{caption}
\usepackage{tabularx}
\usepackage{microtype}
\usepackage{makecell}

\definecolor{cvprblue}{rgb}{0.21,0.49,0.74}
\def\paperID{****} 
\def\confName{CVPR}
\def\confYear{2024}

\title{Generative Denoise Distillation: Simple Stochastic Noises Induce Efficient Knowledge Transfer for Dense Prediction}
\author{Zhaoge Liu$^1$ 
\quad
Xiaohao Xu$^2$ 
\quad
Yunkang Cao$^1$ 
\quad
Weiming Shen*$^1$\\\\
$^1$ Huazhong University of Science and Technology \quad $^2$ University of Michigan
}

\begin{document}
\maketitle
\begin{abstract}
Knowledge distillation is the process of transferring knowledge from a more powerful large model (teacher) to a simpler counterpart (student). Numerous current approaches involve the student imitating the knowledge of the teacher directly. However, redundancy still exists in the learned representations through these prevalent methods, which tend to learn each spatial location's features indiscriminately. To derive a more compact representation (concept feature) from the teacher, inspired by human cognition, we suggest an innovative method, termed Generative Denoise Distillation (GDD), where stochastic noises are added to the concept feature of the student to embed them into the generated instance feature from a shallow network. Then, the generated instance feature is aligned with the knowledge of the instance from the teacher. We extensively experiment with object detection, instance segmentation, and semantic segmentation to demonstrate the versatility and effectiveness of our method. Notably, GDD achieves new state-of-the-art performance in the tasks mentioned above. We have achieved substantial improvements in semantic segmentation by enhancing PspNet and DeepLabV3, both of which are based on ResNet-18, resulting in mIoU scores of 74.67 and 77.69, respectively, surpassing their previous scores of 69.85 and 73.20 on the Cityscapes dataset of 20 categories. The code is available at \href{https://github.com/ZhgLiu/GDD}{https://github.com/ZhgLiu/GDD}.
\end{abstract}
    
\vspace{-1em}
\section{Introduction}
\label{sec:intro}

\begin{figure}[thbp!]
  \centering
  \includegraphics[width=1\linewidth]{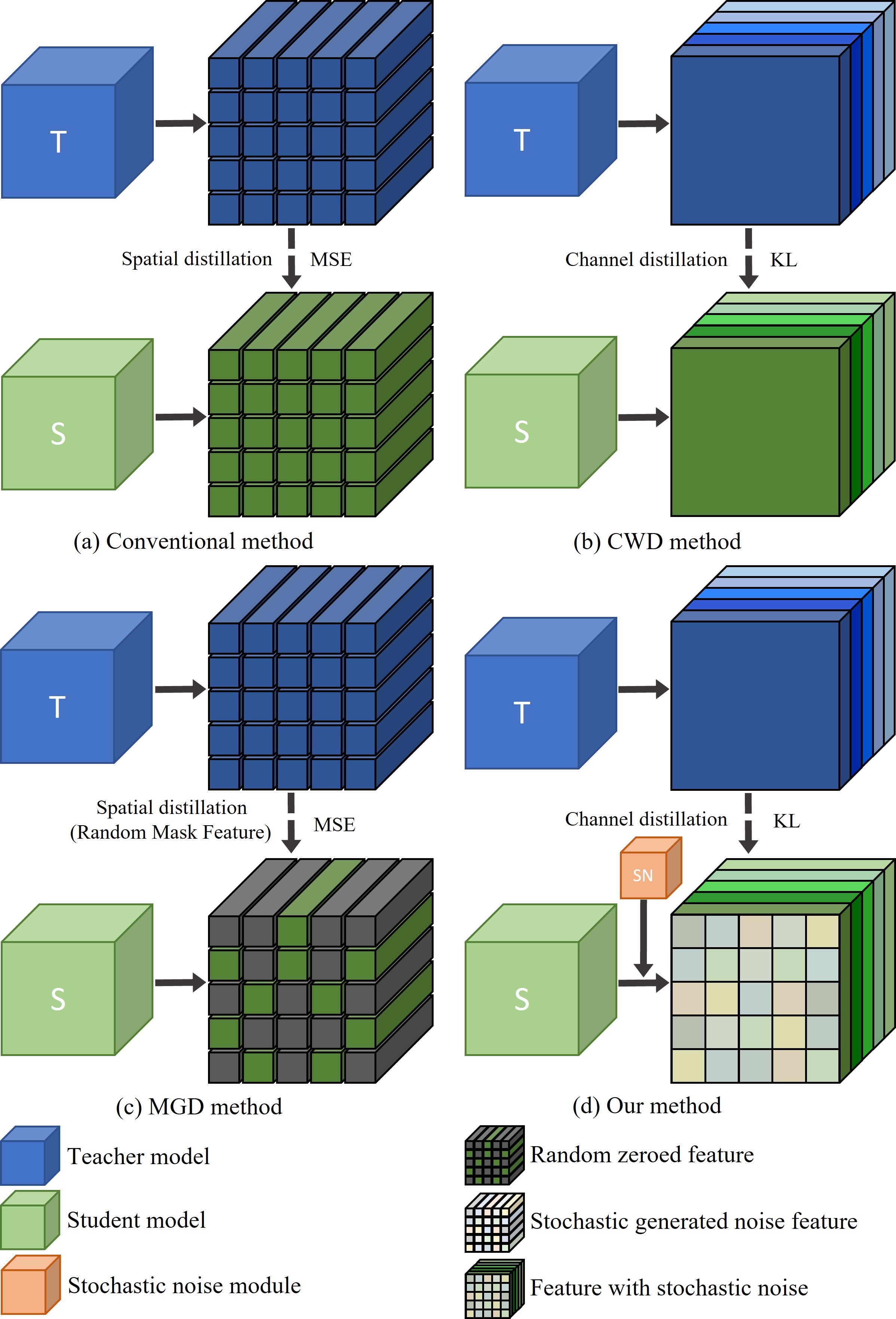}
  \caption{\textbf{Visualization of different knowledge distillation methods.} (a) Conventional spatial knowledge distillation. (b) Channel-wise knowledge distillation. (c) Mask generative knowledge distillation. (d) Stochastic generative knowledge distillation.} 
  \label{compare}
\end{figure}

In recent times, substantial advancements have been made in the realm of large-scale models, exemplified by CLIP~\cite{radford2021learning}, SAM~\cite{kirillov2023segment}, and GPT-4V~\cite{yang2023dawn,GenericAd}. However, the huge dimensions and complexity of these models pose a major challenge for applications with limited computational resources. Within the domain of computer vision, dense prediction~\cite{ranftl2021vision,roland2021rethink} addresses a set of critical issues aimed at establishing connections between input images and intricate output structures. The escalating demand for mobile applications intensifies the hindrance posed by the substantial computational complexity inherent in such tasks. Consequently, recent efforts have pivoted towards the development of neural networks characterized by more compact sizes, reduced computation costs, and superior performance. This study centers specifically on knowledge distillation~\cite{gou2021knowledge, hinton2015distilling}, a methodology that seeks to amplify the performance of a smaller model (student model) by transferring knowledge~\cite{ding2022dual} from a larger model (teacher model).

Knowledge distillation mirrors the pedagogical dynamics inherent in human learning, analogous to the transfer of knowledge from human teachers to students. Explicitly, a teacher assumes the responsibilities of imparting knowledge, teaching skills, and offering guidance and clarification. In turn, students acquire valuable insights from their teachers~\cite{zhu2022teach}, gradually achieving independence and proficiency. In the context of knowledge distillation, the objective is to prompt the student model to attain comparable, if not superior, performance relative to the larger teacher model, all while utilizing fewer resources and demanding less computational power~\cite{yang2020knowledge}. Knowledge distillation has proven to be an effective technique for model compression, finding successful applications across various computer vision tasks, including classification~\cite{zhao2022multi,li2023curriculum,zhu2023bookkd}, detection~\cite{lin2017feature,redmon2016you,li2022knowledge, semi-kd}, and segmentation~\cite{he2017mask,jiang2023masked,CDO}.

In the context of knowledge distillation, teachers often prioritize delivering content directly to students, focusing on minimizing differences between eigenvalues, without adequate consideration for underlying teaching methodologies. Consequently, this approach may lead to suboptimal learning outcomes for students. For example, while feature-based distillation approaches~\cite{IKD,yim2017gift,adriana2015fitnets} have been proposed to enhance student models' performance by emulating the teacher model's embeddings at various spatial locations, as illustrated in Figure~\ref{compare}(a), these methods have demonstrated limited effectiveness in practice. Potential factors contributing to this limitation include dimensional inconsistencies, structural disparities between models, and the risk of overfitting. Mere emulation of the teacher model~\cite{huang2022knowledge} often falls short of significantly enhancing the student model. Thus, it is imperative to incorporate effective learning strategies~\cite{zheng2022localization,zhu2021student} that facilitate knowledge transfer and enhance student model performance. Approaches such as CWD~\cite{shu2021channel} (Figure~\ref{compare}(b)), which transforms the feature dimension of distillation from space to channel, and MGD~\cite{yang2022masked} (Figure~\ref{compare}(c)), which conducts distillation after masking the feature in a patch-wise manner, represent crucial advancements in this direction.

\begin{figure*}[htbp!]
  \centering
  \includegraphics[width=0.9\textwidth]{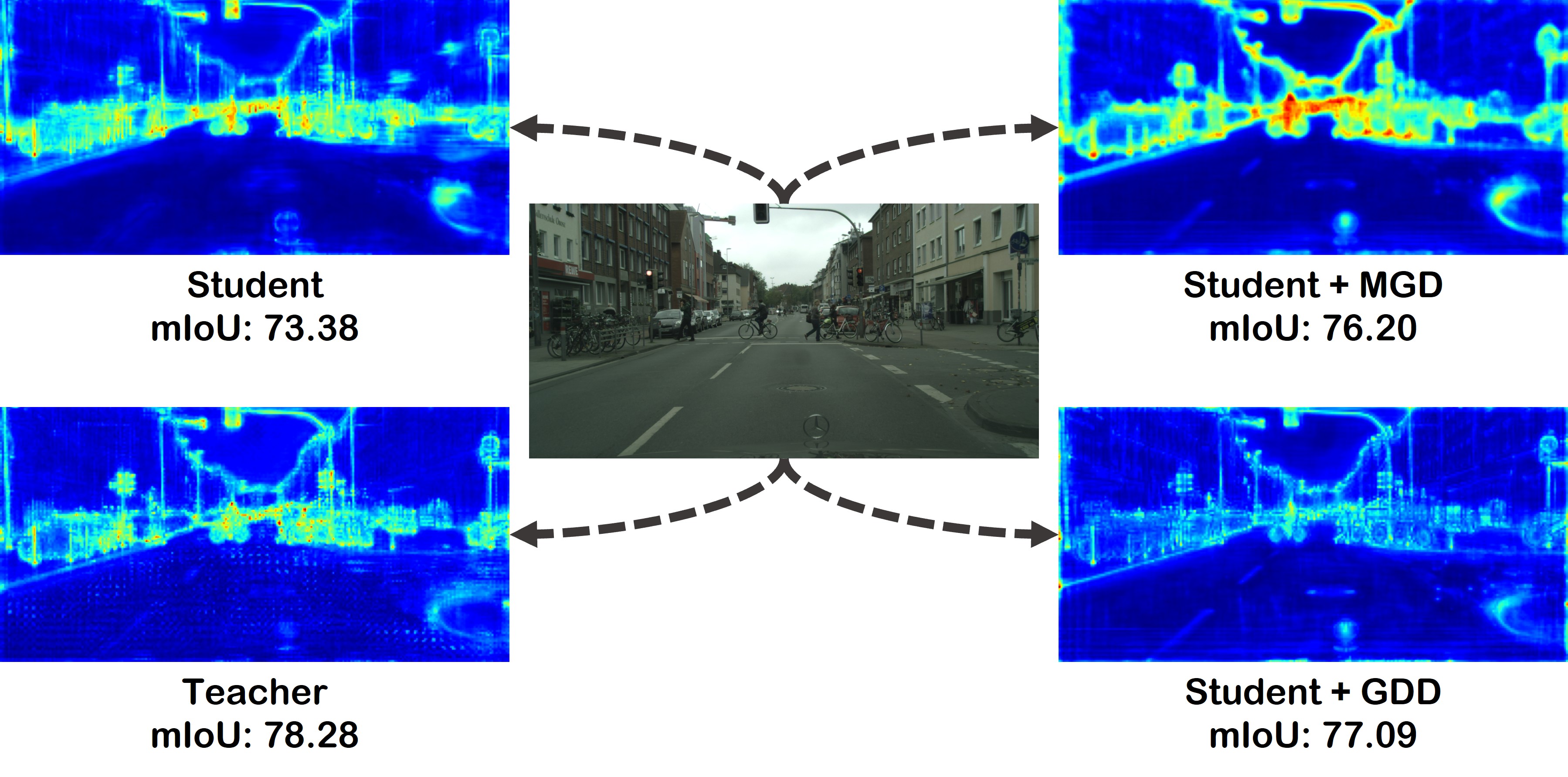}
  \caption{\textbf{Visualization of the backbone layer feature map.} Student: DeepLabV3-Res18, Teacher: PspNet-R101. MGD is one of the most effective distillation methods, whereas GDD is a novel method proposed in this paper.}
  \label{feature}
\end{figure*}

A triumphant individual encounters numerous obstacles and challenges~\cite{jayawickreme2014post}, akin to a student navigating the complexities of the learning process. Whether facing tasks like reading comprehension, multiple-choice assessments, or fill-in-the-blank exercises, the potential for misleading and confounding questions is ever-present. However, it is precisely these challenges that enable students to solidify their understanding of the subject matter. In the context of knowledge distillation, we can apply this concept by introducing artificial perturbations to the student model and leveraging a generation module to enhance the model's expressive capabilities. The inherent vulnerability of students when confronted with intricate inquiries can be mitigated by intentionally infusing the learning process with a controlled level of cognitive noise. This deliberate introduction of perturbations acts as a potent catalyst for enhancing students' adaptability and robustness.

As illustrated in Figure~\ref{compare}(d), our approach involves injecting stochastic noises into the student embedding, which is then processed by a generation module to generate a probability distribution within the channel dimension of the teacher network. Unlike conventional distillation methods that rely on Mean Squared Error for spatial similarity, we employ the KL divergence to characterize distribution consistency. This approach accounts for the possibility that different channels may recognize distinct instance targets within the network. As intuitively observed in Figure~\ref{feature}, our method facilitates a streamlined feature learning process and attains superior performance.

We introduce Generative Denoise Distillation (GDD), an innovative distillation method that incorporates stochastic noise disturbance and channel attention to generate embeddings, thereby enhancing knowledge transfer from teacher to student model. GDD employs diverse noise types, including Gaussian noise, for more efficient knowledge transfer. Unlike conventional approaches that solely emphasize common spatial regions, GDD prioritizes different channels, resulting in a more comprehensive and diverse acquisition of knowledge. Notably, our approach demonstrates state-of-the-art performance in dense prediction tasks.

In summary, the contributions of our paper can be succinctly outlined as follows:
\begin{itemize}
    \item We propose an efficient distillation strategy involving augmenting the student's embedding with Gaussian noise and utilizing a generation module to align it with the teacher's embedding, thereby enhancing the robustness of the student model.
    \item We introduce a channel distillation mechanism that enables the student model to align distinct instances of knowledge with the teacher model effectively. This approach enhances the student model's ability to learn features from each individual object instance more effectively.
    \item We validate the effectiveness of our approach through experiments conducted on COCO and Cityscapes datasets. Across both detection and segmentation tasks, our method achieves extraordinary performance.
\end{itemize}

\section{Related Work}
\label{sec:related}

While a significant body of research on knowledge distillation has been dedicated to overcoming challenges in classification, the growing need for intricate and nuanced solutions remains evident. In response to this demand, our study takes a distinctive approach by aiming to develop sophisticated knowledge transfer techniques specifically crafted to address the complexities of dense prediction tasks. This sets our work apart from conventional feature-based methods.

\subsection{Distillation Algorithm for Image Classification}
Hinton initially proposed knowledge distillation as a technique for improving the performance of classification models~\cite{hinton2015distilling}, in which the output of the teacher model supervises the student model. Furthermore, FitNet~\cite{adriana2015fitnets} leveraged the teacher model's intermediate representations as cues to facilitate the training process and boost the student model's overall performance. DGKD~\cite{son2021densely} employed multiple assistant models to improve the capability of the low-capacity student model, which offers additional guidance throughout the distillation process. DKD~\cite{zhao2022decoupled} reformulated the classical KD loss into two components, allowing the student model to benefit from both accurate and erroneous predictions generated by the teacher model. CTKD~\cite{li2023curriculum} utilized a dynamic and learnable temperature to regulate the level of task difficulty while the student model was being trained. From a holistic perspective, these methods illustrate the efficacy of knowledge distillation in transferring knowledge from high-capacity teacher models to low-capacity student models and emphasize the significance of leveraging the intermediate representations and guidance from multiple sources during the distillation process.

\subsection{Distillation Algorithm for Dense Prediction}
Within the realm of knowledge distillation, there is an ongoing and escalating requirement for dense predictions~\cite{ranftl2021vision,xu2022reliable} of exceptional quality. In contrast to classification, dense prediction tasks present a significantly distinct challenge, requiring prediction not only of category but also of location. To meet these increasing demands, researchers have proposed various feature-based distillation algorithms that aim to improve the fine-grained prediction capabilities of student models. There is a marked shift towards emphasizing feature-level distillation, which has proven to be highly effective in a diverse array of applications. These algorithms, through the transfer of knowledge from the teacher model to the embeddings of the student model, substantially improve the student model's capabilities.

\noindent \textbf{Instance and Semantic Segmentation:}
In the utilization of features, SSTKD~\cite{ji2022structural} framework leveraged both structural and statistical texture knowledge to improve performance. Feature fusion is often used in deep learning, ReviewKD~\cite{chen2021distilling} proposed a cross-level connectivity path that examines connectivity between teacher and student networks at different levels, enhancing feature-based distillation for dense prediction models. Unlike most mean-square error methods for calculating spatial dimensions, CWD~\cite{shu2021channel} transforms channel activations into a probabilistic map and lessens the Kullback-Leibler divergence between teacher and student networks, thereby diminishing the difference between the two models. Instead of learning all the features at once, MGD~\cite{yang2022masked} randomly masked pixels in student features and forces the model to generate the full features of the teacher through a simple block, improving the correspondence of features in dense prediction tasks between teacher and student models.

\noindent \textbf{Object Detection:}
The improvement of knowledge distillation is mainly from the two perspectives of distillation strategy and knowledge selection. For example, GID and FGD both put forward new methods in knowledge selection. In the GID~\cite{dai2021general} approach, the selection of the most distinct instances for distillation from both teacher and student models is adaptively conducted, facilitating a more efficient transfer of knowledge. FGD~\cite{yang2022focal} separated foreground and background information and rebuilt the relationship between pixels to guide student detectors more effectively. Furthermore, novel approaches have been proposed to improve distillation strategies. LD~\cite{zheng2022localization} reformulated the knowledge distillation process for localization and effectively transferred the localization knowledge from the big to the small model.

Contemporary best methods for knowledge distillation in dense prediction tasks frequently rely on mimicking intermediate spatial features~\cite{yang2022masked,chen2021distilling,dai2021general}, which, when directly imposed on the teacher output, can lead to overfitting and suboptimal models. To address this, an efficient knowledge transfer strategy is essential. The challenge is to determine how to learn from the teacher and what to learn. This manuscript introduces a unique method, drawing inspiration from common human learning strategies, which centers on reinforcing the learning process by adding obstacles. Specifically, we employ a channel probability distribution method to select the most prominent instance objects for distillation. By selectively focusing on the most informative instances, we can enhance the effectiveness of knowledge conveyance from the teacher to the student model, resulting in better performance on dense prediction tasks.

\section{Method}
\label{sec:method}

\subsection{Vanilla Knowledge Distillation}

Most knowledge distillation involves two main research directions: logit-based and feature-based distillation. The vanilla method of knowledge distillation was logit-based. 

In the logit-based approach, neural networks calculate the logits $z_i$ for each class, which represent the unnormalized values of the outputs of the network.

Neural networks generally incorporate a "softmax" layer as the output, applying the softmax function to logits to create a probability distribution among the classes. This function compares the logits for each class and normalizes them to produce the probability $p_i$ of each class, which is formulated as follows:
\begin{equation}
    q_i=\frac{e^{p_i/ \mathcal{T}}}{\sum_{j}{e^{p_i / \mathcal{T}}}}
\end{equation}

\noindent Here, the parameter $\mathcal{T}$ represents temperature, which is generally assigned a default value of 1. Increasing $\mathcal{T}$ will soften the class probability distribution.

\noindent The primary logit distillation method can be expressed by the formula:

\begin{equation}
\mathcal{L}_{logit}=D_{KL}\left(q_i^T\ , \ q_i^S\right)
\end{equation}

\noindent Here, $q_i^T$ and $q_i^S$ denote the softmax output of the teacher and student networks, respectively. $D_{KL}\left(\cdot, \cdot\right)$ denotes the Kullback-Leibler divergence, a metric that quantifies the disparity between a pair of probability distributions.

\subsection{Generative Denoise Distillation}

\begin{figure*}[htbp]
  \centering
  \includegraphics[width=0.9\textwidth]{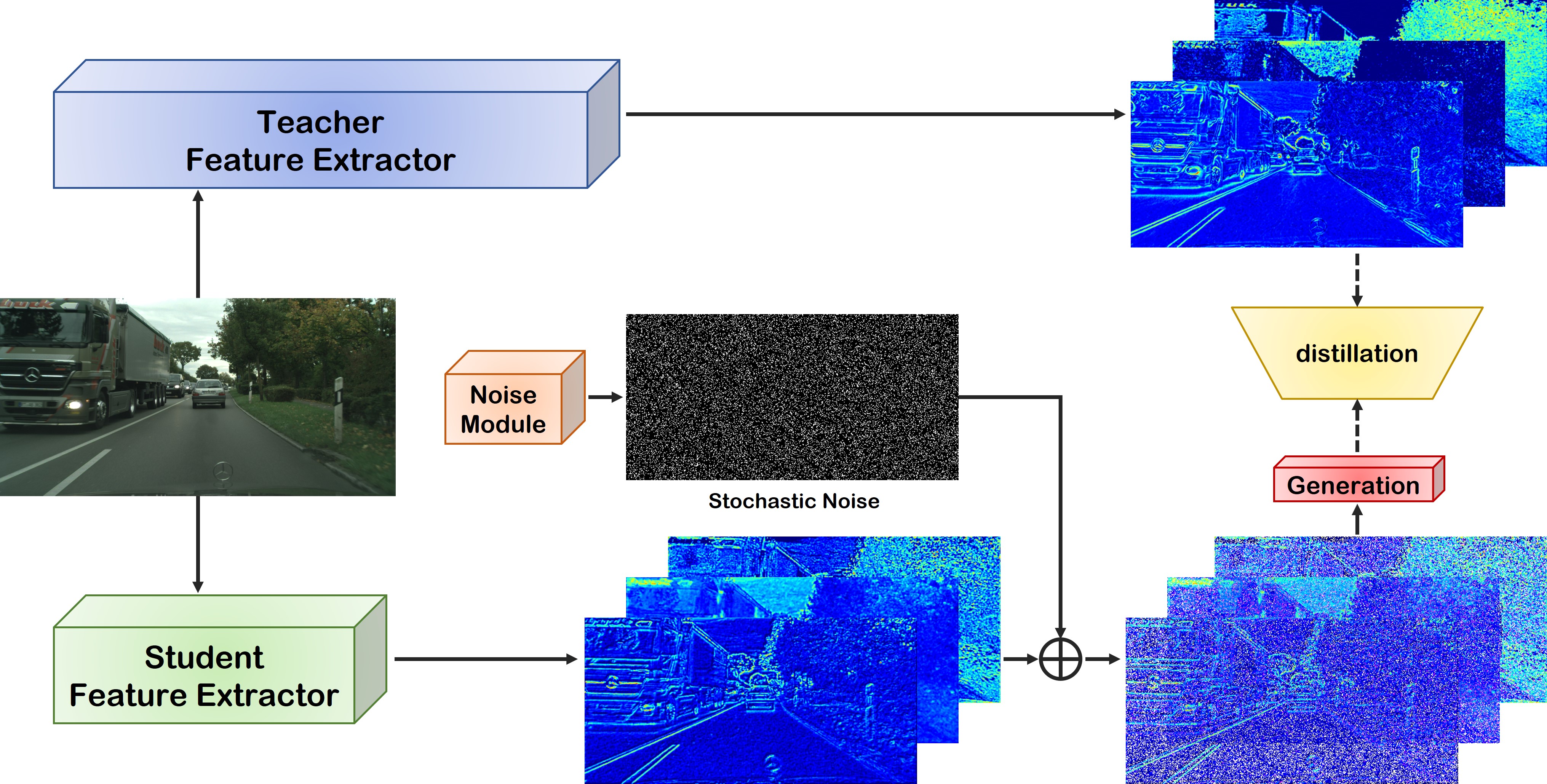}
  \caption{\textbf{Pipeline overview of our method, Generative Denoise Distillation (GDD).} In terms of the student model, we add stochastic Gaussian noise into its feature maps for perturbation, then use a generation module to obtain new feature embedding, and finally distillation for different instance objects in the channel dimension.}
  \label{method}
\end{figure*}

Even in the presence of a blurred image, the recognition of its content remains possible. However, the removal of the blur elevates our perception of the image. Analogously, by introducing perturbations into the student embedding, its ability to assimilate knowledge from the teacher embedding can be improved. To optimize this strategy, it is pivotal to de-blur the embedding and selectively concentrate on distinct instances. This can be achieved by introducing noise to the embedding and employing a generation module for denoising. Furthermore, we can cultivate various channel-embedding probability distributions to augment the student's feature representation capability. The comprehensive architecture of GDD is depicted in Figure \ref{method}, elucidated in subsequent detail.

\subsubsection{Stochastic Noise Generation}

We consider a neural network architecture, specifically focusing on a student network where the concept embedding in an intermediate layer is denoted as $\mathbf{e}$. Under standard conditions with input $\mathbf{x}$ and network parameters $\theta$, this embedding $\mathbf{e}$ manifests as a deterministic entity.
The infusion of stochastic elements into this deterministic system is executed via a corruption function, $c(\cdot,\cdot)$. This function integrates randomness into the concept embedding by superimposing noise, which can be sampled from diverse probabilistic distributions such as normal, Poisson, or Gaussian distributions. This stochastic method diverges significantly from the deterministic perturbation strategies employed in MGD \cite{yang2022masked}, where feature masking is systematic and predictable.

The stochastic noises, particularly sourced from a Gaussian distribution characterized by a mean of zero and a variance of one ($\epsilon \sim \mathcal{N}(0,1)$), introduce an element of variability into the feature representation. This variance is instrumental in the process of feature perturbation, enhancing the complexity and potentially the efficacy of the neural network's learning process.

\subsubsection{Instantiation Denoise Network}
To begin the process, we add Gaussian noise to cover the k-th feature of the student, formulated as follows:

\begin{equation}
    G(z)=\frac{1}{\sigma\sqrt{2\pi}}e^{-{(z-\mu)}^2/\left(2\sigma^2\right)}
\end{equation}

Here $G_z$ represents random Gaussian noise, with $\mu=0$, $\sigma=1$.

To simplify notation, we will use $T$ to refer to the activation maps of the teacher and $S$ to refer to those of the student. In particular, the symbols $T_c\in \mathbb{R}^{C\times H\times W}$ and $S_c\in \mathbb{R}^{C\times H\times W}$ (where $(c=1, \ldots, C)$) denote the c-th channel's feature map for the teacher and student models, respectively.

The $A_{lign}$ adaptation layer employs $1\times1$ kernel convolutional layers. The c-th noisy feature map $G(z_c)$ is generated by introducing Gaussian noise into the corresponding feature map. Subsequently, we attempt to produce the teacher's clean feature map by utilizing the respective noisy feature map derived from the student. This process can be expressed as follows:
\begin{equation}
    S'_c=G_e\left(\ A_{lign}\left(S_c\right)+G(z_c)\right)\rightarrow T_c
\end{equation}
\begin{equation}
    G_e\left(x\right)={Conv}_{l2}(ReLU({Conv}_{l1}(x)))
\end{equation}
$G_e$ represents the projector layer, comprising two $3\times3$ convolutional layers (${Conv}_{l2}$ and ${Conv}_{l1}$) and a $ReLU$ activation layer.

\subsubsection{Channel Knowledge Alignment}
The loss associated with channel dimension distillation can be articulated as $D\left(\phi\left(T\right),\phi\left(S\right)\right)$, this loss quantifies the dissimilarity between $\phi\left(T\right)$ and $\phi\left(S\right)$ in terms of their channel dimensions.

In this scenario, the function $\phi(\cdot)$ is employed for transforming values of activation into a corresponding probability distribution, detailed as follows:

\begin{equation}
    \phi(x_{c,i}) = \frac{e^{\left(\frac{x_{c,i}}{\mathcal{T}}\right)}}{\sum_{k=1}^{HW} e^{\left(\frac{x_{c,i}}{\mathcal{T}}\right)}}
\end{equation}

The $D_{KL}(\cdot)$ function is used to assess the differences in channel distributions between the teacher and student networks. The index $i$ refers to the spatial location of the current channel.
\begin{equation}
    D_{KL}\left(x^T,x^S\right) = \frac{\mathcal{T}^2}{C} \sum_{c=1}^{\mathcal{C}} \sum_{i=1, j=1}^{H \times W} \phi({x_{c,i}^T}) \log \frac{\phi({x_{c,i}^T})}{\varphi({x_{c,i}^S})}
\end{equation}

Based on the approach we have described, we design the distillation $\mathcal{L}_{distill}$ for GDD as follows:
\begin{equation}
    \mathcal{L}_{distill}\left(x^T,x^S\right)=D_{KL}\left(T, \ S'\right)\ 
\end{equation}
Here, $x^T$ and $x^S$ denote the parameters of the teacher and academic model that are ultimately involved in the loss calculation, respectively.

\subsubsection{Overall Loss}
Our model's cumulative loss function consists of two key elements: the task-oriented loss, denoted as $\mathcal{L}_{task}$, and the knowledge distillation loss, represented by $\mathcal{L}_{distill}$. The loss associated with the task gauges the model's efficacy in a particular downstream application, whereas the distillation loss facilitates the transference of knowledge from a larger, intricate teacher model to a more compact and straightforward student model.
The two components are combined using a weighting factor $\alpha$, culminating in the establishment of the overall loss equation $\mathcal{L}$ as follows:

\begin{equation}
    \mathcal{L} = \mathcal{L}_{task}+\alpha\cdot \mathcal{L}_{distill}
\end{equation}

GDD is a straightforward and efficient approach to distillation that is easily adaptable across various tasks. The implementation procedure of this method is concisely outlined in Algorithm \ref{algo}.

\begin{algorithm}[H]
\caption{Generative Denoise Distillation}
\label{algo}
\begin{enumerate}
    \item \textbf{Input:} Data: $x$, Label: $y$, Teacher model: $Tea$, Student model: $Stu$, Align module: $A_{lign}$, Gaussian distribution noise: $G(z)$, Generation module: $G_{e}$, Hyper-parameter: $\alpha$
    \item Obtain $S$ and $\hat{y}$ by passing input $x$ through $Stu$.
    \item Obtain $T$ by passing input $x$ through $Tea$.
    \item Align channel dimensions of $S$ with $T$ using $A_{lign}$ if needed.
    \item Add Gaussian noise $G(z)$ to aligned $A_{lign}(S)$, resulting in noisy embedding: $A_{lign}(S)+G(z)$, where $G(z)$ follows a Gaussian distribution with $\mu=0$ and $\sigma=1$.
    \item Using Generation module $G_{e}$ to get a new student embedding $S'=G_{e}(A_{lign}(S)+G(z))$.
    \item Calculate distillation loss $\mathcal{L}_{distill}(T,S')$ using embeddings $S'$ and $T$.
    \item Calculate task loss $\mathcal{L}_{task}(\hat{y},y)$ using output $\hat{y}$ and label $y$.
    \item Calculate overall loss $Loss=\mathcal{L}_{task}+\alpha\cdot \mathcal{L}_{distill}$ for updating $Stu$.
    \item \textbf{Output:} Updated Student model: $Stu$
\end{enumerate}
\end{algorithm}

\section{Experiments}
\label{sec:experiments}
In the current section, we assess how effective our GDD, a feature-based distillation approach, is across various tasks including semantic segmentation, instance segmentation, and object detection. The experimental outcomes show remarkable performance across all these approaches when utilizing GDD.

\subsection{Datasets}

\textbf{Cityscapes:}
Designed explicitly for understanding urban scenes semantically, the Cityscapes dataset includes 5,000 images, each carefully annotated. Of these, 2,975 are designated for training purposes, 500 serve as validation, and 1,525 are used for testing. Among the 30 usual classes represented in these images, 19 are utilized specifically for evaluation and testing purposes. Each image measures $2048 \times 1024$ pixels and has been collected from 50 distinct cities. In our experiments, we focus solely on the finely labeled data, excluding the coarsely labeled portion.

\noindent \textbf{MS COCO2017:}
The COCO (Common Objects in Context) dataset is a substantial resource tailored specifically for tasks such as object detection, segmentation, and captioning. It comprises over 200,000 meticulously annotated images, with 118,000 allocated for training, and 5,000 for validation. Within these images, 91 common object types are grouped into 80 different object categories used for evaluation. Each image, collected from a variety of scenes, typically contains multiple objects.

\subsection{Evaluation Metrics}

\textbf{Semantic Segmentation:}
For evaluating the efficacy and output of our dense prediction-focused distillation method, we follow previous studies by testing each strategy through the mean Intersection-over-Union (mIoU), which evaluates the overlap of predicted and actual ground-truth masks for each individual instance across all single-scale setting experiments.

\noindent \textbf{Instance Segmentation and Object Detection:}
For measuring instance segmentation performance, we leverage the mean Average Precision (mAP) score. This metric evaluates the precision of identified object instances, calculating the overlap between the predicted bounding boxes and masks and their corresponding ground-truth counterparts.

For instance segmentation and object detection, model performance evaluation extends to objects of varying sizes. To address this, researchers commonly employ specific metrics, namely AP${}_{L}$, AP${}_{M}$, and AP${}_{S}$, representing average precision for large, medium, and small objects, respectively.

\subsection{Implementation Details}
We conducted experiments on a system with 8 Tesla V100 GPUs to evaluate the effectiveness of knowledge distillation within the PyTorch framework. The software setup included PyTorch 1.10.1, CUDA 11.1, and torchvision 0.11.2. During training, we employed a batch size of 16. The optimization was performed using the SGD optimizer. If not stated otherwise, We subject the models to a training regimen of 24 epochs (12 epochs for SOLO)  and set the distillation temperature parameter $\tau$ to 4 following CWD \cite{shu2021channel}.

In semantic segmentation, our approach uses an SGD optimizer that features a 0.9 momentum and sets the weight decay at 0.0005, with experiments carried out using the MMSegmentation framework~\footnote{MMSegmentation is an open source semantic segmentation toolbox based on PyTorch.}. 

In instance segmentation and object detection, we similarly employ the SGD optimizer, but adjust the weight decay to 0.0001, conducting experiments via the MMDetection framework~\footnote{MMDetection is an open source object detection toolbox based on PyTorch.}. During training, we adopt an inheriting strategy \cite{yang2022focal, kang2021instance} where the student model inherits both neck and head parameters from the teacher if they have identical head structures.

\subsection{Main Results}
We conducted a thorough comparison of the inference results for the three major dense prediction tasks in computer vision. The visualizations unequivocally indicate that our proposed GDD method outperforms others in terms of performance and accuracy.

\subsubsection{Semantic Segmentation}

\begin{figure}[htbp!]
  \centering
  \includegraphics[width=1\linewidth]{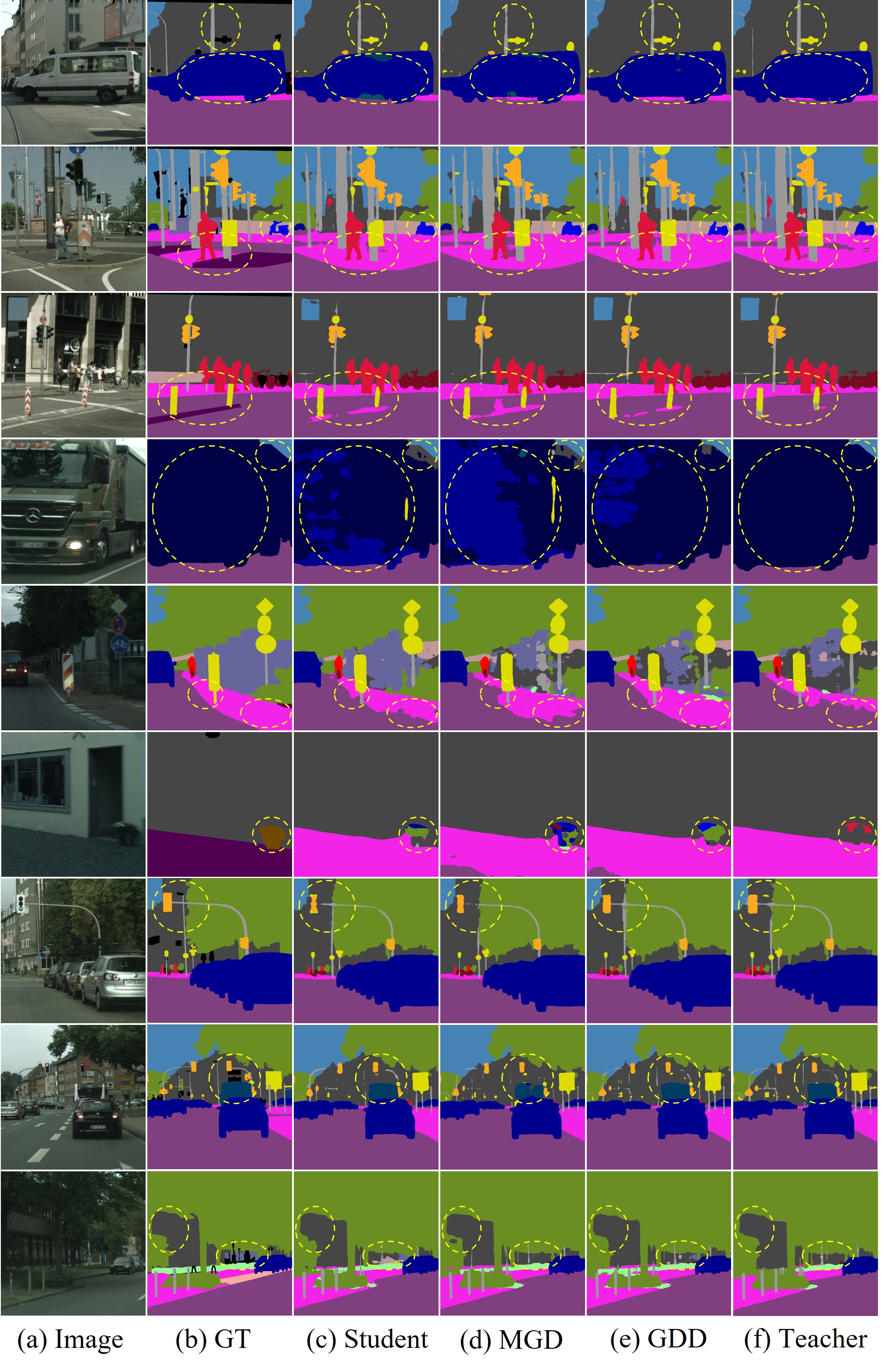}
  \caption{\textbf{Semantic segmentation: qualitative comparative results.} (a) Image: raw image, (b) GT: ground truth, (c) Student: DeepLabV3-Res18, (d) Mask Generative distillation \cite{yang2022masked}, (e) GDD: Generative Denoise Distillation, (f) Teacher: PspNet-Res101.}
  \label{semantic}
\end{figure}

\begin{table}[ht!]
\caption{Semantic segmentation performance on the cityScapes dataset. {T} symbolizes the teacher network, while {S} represents the student network.}
\label{city-semantic}
\centering
\begin{tabular}{@{}lcc@{}}
\toprule
Method & Input Size  & mIoU\\ 
\midrule
PspNet-Res101 (T)&$512\times 1024$&78.34\\ \hline
DeepLabV3-Res18 (S) & $512\times 512$ &73.20\\
+SKDS &$512\times 512$&73.87\\
+CWD &$512\times 512$&75.93\\   +MGD&$512\times 512$&76.02\\
+\textbf{GDD} & $512\times 512$&{\bf77.09}\\
\hline \hline
PspNet-Res101 (T) & $512\times 1024$ & 78.34\\ \hline
PspNet-Res18 (S) & $512\times 512$ & 69.85\\
+SKDS & $512\times 512$ & 72.70\\
+CWD & $512\times 512$ & 73.53\\
+MGD & $512\times 512$ & 73.63\\
+\textbf{GDD} & $512\times 512$&{\bf74.67}\\
\bottomrule
\end{tabular}
\end{table}

\noindent \textbf{Details of distillation:}  
Evaluation of the distillation loss occurs using feature maps derived from the final feature map in the backbone. The hyper-parameters determined for each model variant are as follows: PspNet with $\alpha$ = 145 and DeepLabV3 with $\alpha$ = 175. 

\begin{table}[ht!]
\caption{Evaluation of GDD versus another cutting edge knowledge distillation method in class-level IoU on Cityscapes validation set, featuring DeeplabV3-Res18 (student network) and PspNet-Res101 (teacher network).}
\label{city-class}
\centering
\resizebox{\linewidth}{!}{
\begin{tabular}{@{}c|ccccc@{}}
\toprule
Method & fence & pole &traffic light & traffic sign & vegetation\\ \hline
+MGD &58.45 &63.64  &70.54  &78.27  &91.93 \\
\textbf{+GDD}  &{\bf59.6} &{\bf63.76} &{\bf71.54} &{\bf78.66} &{\bf92.13} \\\midrule
Method & bicycle & road & sidewalk & building &  wall\\ \hline
+MGD & 77.33      &{\bf98.03}  &{\bf84.37}   &91.97  &44.0\\
\textbf{+GDD} & {\bf77.67}    &97.95 & 84.18 &{\bf92.13} &{\bf48.06}\\
\midrule
Method & terrain & sky &person &rider & car\\ \hline
+MGD &63.37 &94.23 &81.30 &{\bf61.31} &94.52\\
\textbf{+GDD}  &{\bf64.33}  &{\bf94.45} &{\bf81.52}  &60.91 &{\bf94.80}\\
\midrule 
Method & truck  &bus &train & motorcycle &    \multicolumn{1}{|c}{\textbf{mIoU}}\\ \hline
+MGD & 69.02 &87.39 &{\bf77.55} &60.55 &\multicolumn{1}{|c}{76.20} \\
\textbf{+GDD} &{\bf74.40} &{\bf88.20} &76.62 &{\bf63.79} & \multicolumn{1}{|c}{\bf77.09}\\
\bottomrule
\end{tabular}}
\end{table}

\noindent \textbf{Results:}
In this study focusing on semantic segmentation, we explore two experimental settings. For both, the PspNet-Res101 \cite{zhao2017pyramid} is utilized as the teacher model, undergoing training for 80,000 iterations with an input resolution of 512×1024. As student models, we deploy PspNet-Res18 and DeepLabV3-Res18 \cite{chen2017rethinking}, each trained for 40,000 iterations using the same input dimensions. As indicated in Table \ref{city-semantic}, our approach outperforms the leading distillation methods in semantic segmentation. This superiority is also visually evident in Figure \ref{semantic}. Our distillation techniques, both homogeneous and heterogeneous, yield substantial enhancements for the student models. Notably, the PspNet-Res18 achieves an increase of 4.82 mIoU.

Additionally, Table \ref{city-class} presents a detailed comparison of class IoU between our method and the latest top-performing approach, MGD \cite{yang2022masked}. We observe notable enhancements in class accuracy for various objects including walls, fences, traffic lights, the sky, trucks, and motorcycles with our methods., indicating that the addition of stochastic noises can well transfer important knowledge.

\subsubsection{Instance Segmentation}

\begin{table}[H]
\caption{Instance segmentation performance on the COCO dataset. {\bf MS} denotes training at multiple scales. AP in this scenario refers specifically to Mask Average Precision.}
\label{coco-instance}
\centering
\resizebox{\linewidth}{!}{
\begin{tabular}{@{}clcccc@{}}
\toprule
Teacher & Student & mAP  & AP$_{S}$ & AP$_{M}$ &AP$_{L}$\\ 
\midrule
\multirow{3}{*}{\makecell{SOLO-Res101, MS\\
(mAP 37.1)}}
    &SOLO-Res50 & 33.1&12.2&36.1&50.8\\
    &+MGD & 36.2&14.2&39.7&{\bf55.3}\\
    &\textbf{+GDD} & {\bf36.4}&{\bf14.6}&{\bf40.1}&55.0\\
\midrule
\multirow{3}{*}{\makecell{SparseInst-Res50\\
(mAP 33.4)}}
    &SparseInst-Res18 & 29.8 & 11.6 & 30.7 & 46.4\\
    &+MGD & 30.9 & {\bf12.4} & 32.1 & 48.1\\
    &\textbf{+GDD} & {\bf31.6}&11.8&{\bf32.9}&{\bf49.3}\\
\bottomrule
\end{tabular}}
\end{table}

\textbf{Details of distillation:} 
Evaluation of the distillation loss takes place on feature maps, which are derived from the models' neck region. The hyper-parameters determined for each model variant are as follows: SOLO with $\alpha$ = 200 and SparseInst with $\alpha$ = 65. 

\noindent \textbf{Results:}
In our instance segmentation study, we experiment with two distinct types of detectors: A one-stage detector (SOLO), and a DETR paradigm detector (SparseInst). In Table \ref{coco-instance}, the effectiveness of GDD is compared with top distillation approaches for detectors, demonstrating superior performance over SparseInst and SOLO.

\subsubsection{Object Detection}

\textbf{Details of distillation:} Evaluation of the distillation loss takes place on feature maps, which are derived from the models' neck region. The hyper-parameters determined for each model variant are as follows: Cascade Mask RCNN with $\alpha$ = 4 and Reppoints with $\alpha$ = 65.

\begin{table}[htbp!]
\caption{Object detection performance on the COCO dataset}
\label{coco-object}
\centering
\resizebox{\linewidth}{!}{
\begin{tabular}{@{}clcccc@{}}
\toprule
Teacher & Student & mAP  & AP$_{S}$ & AP$_{M}$ &AP$_{L}$\\ 
\midrule
\multirow{6}{*}{\makecell{Cascade\\Mask RCNN\\ResNeXt101\\(mAP 47.3)}}
    &Faster RCNN-Res50 & 38.4 &21.5&42.1&50.3\\
    &+FKD & 41.5 & 23.5 & 45.0 & 55.3\\
    &+CWD & 41.7 & 23.3 & 45.5 & 55.5\\
    &+FGD & 42.0 & 23.9 & 46.4 & 55.5\\
    &+MGD & 42.1 & {\bf23.7} & {\bf46.4} & 56.1\\
    &+\textbf{GDD} & {\bf42.2}&23.4&45.8&{\bf56.4}\\
\midrule
\multirow{6}{*}{\makecell{RepPoints\\ResNeXt101\\(mAP 44.2)}}
    &RepPoints-Res50 & 38.6& 22.5 & 42.2 & 50.4\\
    &+FKD & 40.6 & 23.4 & 44.6 & 53.0\\
    &+CWD & 42.0 & 24.1 & 46.1 & 55.0\\
    &+FGD & 42.0 & 24.0 & 45.7 & 55.6\\
    &+MGD & 42.3 & {\bf24.4} & 46.2 & 55.9\\
    &+\textbf{GDD} & {\bf42.7} & 24.3 & {\bf46.8} & {\bf56.8}\\
\bottomrule
\end{tabular}}
\end{table} 

\begin{figure}[thbp!]
  \centering
\includegraphics[width=1\linewidth]{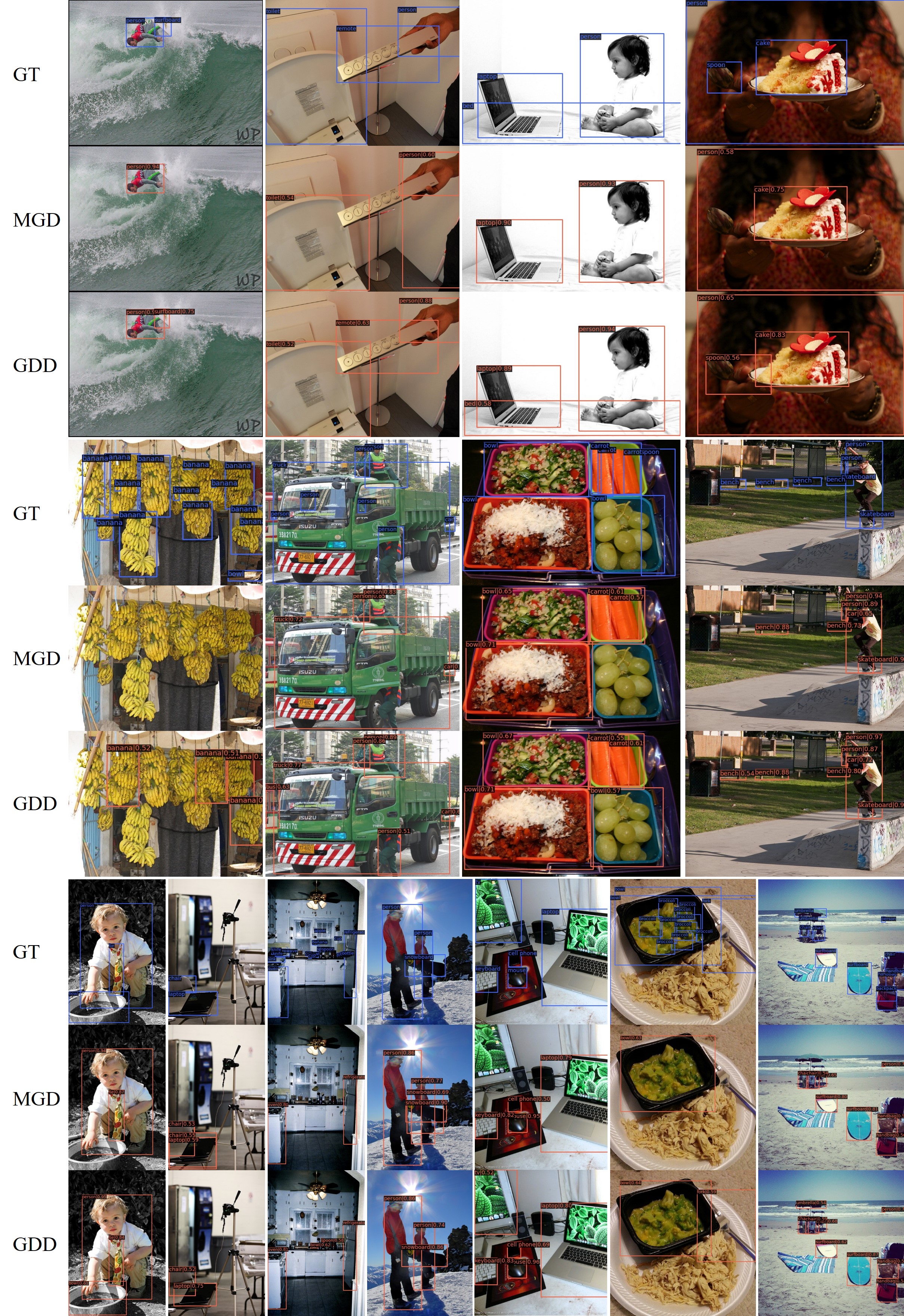}
  \caption{\textbf{Object detection: qualitative comparative results.} GT: ground truth, MGD: Mask Generative distillation \cite{yang2022masked}, GDD: Generative Denoise Distillation.}
  \label{object}
\end{figure}

\noindent \textbf{Results:} In our object detection research, we analyze GDD's performance against several detector distillation methods, shown in Table \ref{coco-object}. GDD notably outperforms others, especially with the RepPoints method, improving 4.1 above the baseline and 0.4 over the prior best method. The predicted results comparison is in Figure \ref{object}.

\section{Analysis}
\label{sec:analysis}

\subsection{Ablation Study of Different Module}
GDD, our method, primarily incorporates two modules: Stochastic Noise (SN) and Channel Distillation (CD). These modules serve distinct roles within the overall method. In the current section, we present an analysis of the roles played by these two modules based on experimental results.

\begin{table}[htbp]
\caption{Component Effectiveness Study.}
\label{CD-SN}
\centering
\begin{tabular}{ccc}
\toprule
Model  & Setting & mAP \\
\midrule
\multirow{4}{*}{\centering DeepLabV3-Res18}
& Baseline &  73.20  \\
& +CD & 75.98   \\
& +SN & {76.46}   \\
& +CD\&SN & \textbf{77.09}   \\
\bottomrule
\end{tabular}
\end{table}

As depicted in Table \ref{CD-SN}, both the CD and SN modules exhibit gains in enhancing model performance. Notably, the latter module demonstrates a particularly significant improvement, and the synergistic effect of integrating both modules is worth highlighting. This ablation experiment effectively demonstrates the efficacy of the distinct components within the proposed method.

\subsection{Effect of Different Noise Strength}

\begin{table}[htbp]
\caption{Results of the effect of different noise strengths for RepPoints-Res50.}
\label{strength-reppoints}
\centering
\begin{tabular}{c|l|c}
\toprule
Model  & Setting & mAP \\
\midrule
\multirow{5}{*}{\makecell{RepPoints-Res50\\+GDD}}
&$\mu=0$, $\sigma=0$& 42.4\\
&$\mu=0$, $\sigma=0.5$& 42.6\\
&$\mu=0$, $\sigma=1$&{\bfseries 42.7}\\
&$\mu=0$, $\sigma=1.5$& 42.5\\
&$\mu=0$, $\sigma=2$& 42.4\\
\bottomrule
\end{tabular}
\end{table}

The examination of how varying noise intensities impact the performance of knowledge distillation indicated that the ideal noise level is contingent on the specific task and dataset.

\begin{table}[htbp!]
\caption{Results of the effect of different noise strengths for DeepLabV3-Res18.}
\label{strength-deeplabv3}
\centering
\begin{tabular}{c|l|c}
\toprule
Model & Setting & mIoU \\
\midrule
\multirow{5}{*}{\makecell{DeepLabV3-Res18\\+GDD}}
&$\mu=0$, $\sigma=0$& 76.12\\
&$\mu=0$, $\sigma=0.5$& 76.68\\
&$\mu=0$, $\sigma=1$&{\bf77.09}\\
&$\mu=0$, $\sigma=1.5$& 76.43\\
&$\mu=0$, $\sigma=2$& 75.94\\
\bottomrule
\end{tabular}
\end{table}

We tested a range of noise strengths on a detection task using our proposed GDD method and found that low levels of noise strength resulted in poor knowledge transfer, while high levels of noise strength resulted in overly noisy generated instances. However, we found that moderate levels of noise strength produced the best knowledge distillation performance, as evidenced by the student model's attainment of RepPoints and DeepLabV3 accuracy on the COCO and Cityscapes datasets. This observation is elucidated in Table \ref{strength-reppoints} and Table \ref{strength-deeplabv3}. These findings have important implications for future research, suggesting that more complex or adaptive methods for adjusting noise strength may be required to achieve optimal performance on different tasks and domains.

\subsection{Effect of Different Noise Injection Location}

In Table \ref{inject-location}, our analysis of the effect of different noise injection locations on the quality of the generated instances revealed that injecting noise in the hidden latent feature produces much better results than injecting noise in the original image. We tested a range of noise injection locations on the segmentation task using our proposed GDD method and found that noise injection in the original image resulted in a worse performance with an accuracy of only RepPoints on COCO dataset. On the other hand, noise injection in the hidden latent feature brings about better performance.

\begin{table}[H]
\caption{Results of the effect of different noise injection locations for RepPoints-Res18}
\label{inject-location}
\centering
\begin{tabular}{ccc}
\toprule
Model  & Inject Location & mAP \\
\midrule
\multirow{2}{*}{\centering RepPoints-Res18}
& Image   & 33.4\\
& Feature & {\bf42.7}\\
\bottomrule
\end{tabular}
\end{table}

The reasons for this outcome can be attributed to the following factors. Firstly, the introduction of noise into the original image may introduce irrelevant or misleading information that hampers the segmentation task. Conversely, injecting noise into the hidden latent feature enables the model to focus on the intrinsic structure and semantics of the image. This process of noise filtering assists the model in generating more accurate segmentation predictions. Moreover, the hidden latent feature space often exhibits nonlinear relationships with the original image space. Injecting noise directly into the original image has the potential to disrupt these nonlinear relationships, making it more challenging for the model to learn meaningful representations. In contrast, injecting noise into the hidden latent feature space, designed to capture relevant information, allows the model to better adapt to the noise and preserve the integrity of the learned representations.

\subsection{Visualization and Qualitative Comparison}

\begin{figure}[h]
  \centering
  \includegraphics[width=1\linewidth]{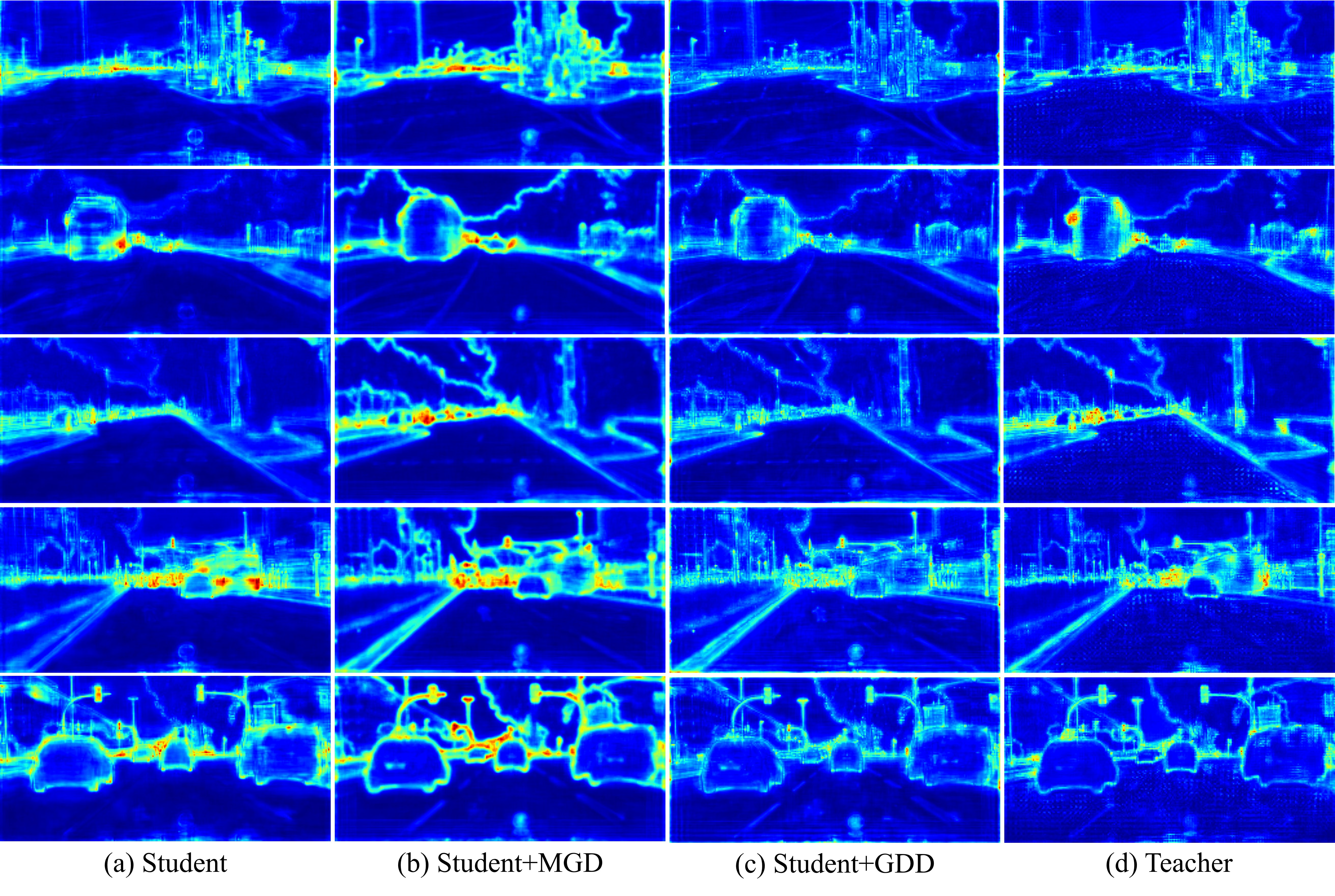}
  \caption{\textbf{Qualitative comparison of the backbone layer feature map.} (a) Student: DeepLabV3-Res18, (b) Student+MGD \cite{yang2022masked}, (c) Student+GDD, (d) Teacher: PspNet-Res101.}
  \label{cityfeature}
\end{figure}

As shown in Figure \ref{cityfeature}, we compared the feature maps generated by our method with those generated by another popular knowledge distillation methods on a detection task using a PspNet-Res101 architecture and DeepLabV3-Res18 architecture for teacher and student models, respectively. Our method generated more distinct and meaningful features, with clear boundaries between objects and backgrounds, thereby advancing toward the attainment of a more robust teacher model. In contrast, the other method generated more noisy and ambiguous features, overlapping boundaries between different objects. These findings have important implications for future research, suggesting that the specific features being learned and transferred between the teacher and student models are critical for achieving optimal performance in knowledge distillation. The qualitative visualization provides valuable insights into these features and how they can be improved.
\section{Conclusion and Discussion}
\label{sec:conclusion}

\subsection{Conclusion}

The proposed method of Generative Denoise Distillation (GDD) represents an innovative approach to knowledge distillation. By promoting the learning of a compact concise representation of instances in the student model, GDD demonstrates greater efficiency and effectiveness in knowledge transfer from the teacher model. Introducing simple stochastic noises into the student's concept feature, and its subsequent embedding into the generated instance feature, aligns well with human cognition. The validation of the effectiveness of GDD on various tasks, including detection and segmentation for generic scenarios, highlights its potential for real-world applications. The achievement of new state-of-the-art results in semantic segmentation tasks further underscores the significance of this novel approach to knowledge distillation.

\subsection{Potential Bias}
In machine learning, addressing bias in models is of paramount importance, as it can lead to inaccurate or unfair predictions. This issue is particularly problematic in knowledge distillation, where the student model is designed to learn from the teacher model and may inherit biases. To mitigate this risk, a potential solution is to use multiple teacher models and incorporate adversarial training. This approach can ensure that the student model learns numerous kinds of features and is less likely to rely on biases from any one teacher model. The potential benefits of this approach include more accurate and fair predictions, as well as increased robustness to variations in the data.

\subsection{Future Work}
In this study, we introduce a straightforward Generative Denoise Distillation method that adds simple stochastic noises to the concept feature of the student model. Future research could investigate the effect of different types of noise on the knowledge transfer process. For example, adding more complex noises or even adversarial noise could potentially improve the quality of the generated instances and enhance knowledge transfer. Besides, our proposed method has been validated on dense prediction tasks, encompassing both detection \cite{carion2020end} and segmentation \cite{he2017mask}. Future research could investigate the impact of noise on more diverse tasks and domains, such as natural language processing \cite{devlin2018bert}. This could help to identify the types of tasks where noise-based knowledge distillation is most effective. What's more, the proposed method adds noise to the concept feature of the student model to encourage it to acquire a more compact representation of instances. Future research could investigate how to incorporate feedback from the student model throughout the process of knowledge transmission. This could potentially enable the student model to provide feedback on the quality of the generated instances and improve the knowledge transfer.
 
{\small
\bibliographystyle{unsrt}
\bibliography{main}
}

\end{document}